\documentclass[10pt,journal]{IEEEtran}

\usepackage{amsmath}
\usepackage{amssymb}
\usepackage{amsthm}
\usepackage{algorithm}
\usepackage{tikz}
\usepackage{xifthen}
\usepackage{tabu}
\usepackage{graphics}
\usepackage{subfig}
\usepackage{epstopdf}
\usepackage{epsfig}
\usepackage{caption}
\usepackage{float}
\usepackage{pgfplotstable}
\usepackage{pgfplots}
\newtheorem{theorem}{Theorem}
\newtheorem{lemma}[theorem]{Lemma}         

\newtheorem{definition}[theorem]{Definition}

\def\x{{\mathbf x}}
\def\R{{\mathbb{R}}}

\newcommand{\norm}[1]{\|#1\|}

\def\1{\mathbf{1}}     
\usepackage{enumitem}

\usepackage{todonotes}

\usepackage{algpseudocode}
\usepackage{cite}
\usepackage{amsfonts}
\usepackage{mathtools}
\usepackage{microtype}
\usepackage{hyperref}

\DeclareMathOperator*{\argmin}{arg\,min}

\def\R{{\mathbb{R}}}
\newcommand{\mb}[1]{\mathbf{#1}}
\def\sgn{\mathrm{sgn}}
\def\card{\mathrm{card}}
\def\t{\intercal}
\def\t{\top}

\newcommand{\eps}{\epsilon}

\usepackage{hyperref}
\usepackage[utf8]{inputenc}
\usepackage{booktabs}
\usetikzlibrary{calc}
\begin{document}
\title{Signal Reconstruction from Modulo Observations}

\author{
	Viraj Shah and Chinmay Hegde \\
	Electrical and Computer Engineering Department \\
	Iowa State University \\
	Ames, IA, USA 50010
	\thanks{Email: \{viraj,chinmay@iastate.edu\}. This work was supported by grants CCF-1566281 and CAREER CCF-1750920 from the National Science Foundation, a faculty fellowship grant from the Black and Veatch Foundation, and a GPU grant from the NVIDIA Corporation.	
	The authors thank Praneeth Narayanamurthy, Gauri Jagatap, and Thanh Nguyen for helpful comments.
}
}

\maketitle

\begin{abstract}
	We consider the problem of reconstructing a signal from under-determined \emph{modulo} observations (or measurements). This observation model is inspired by a (relatively) less well-known imaging mechanism called modulo imaging, which can be used to extend the dynamic range of imaging systems; variations of this model have also been studied under the category of \emph{phase unwrapping}. Signal reconstruction in the under-determined regime with modulo observations is a challenging ill-posed problem, and existing reconstruction methods cannot be used directly. In this paper, we propose a novel approach to solving the inverse problem limited to two modulo periods, inspired by recent advances in algorithms for \emph{phase retrieval} under sparsity constraints. We show that given a sufficient number of measurements, our algorithm perfectly recovers the underlying signal and provides improved performance over other existing algorithms. We also provide experiments validating our approach on both synthetic and real data to depict its superior performance.
\end{abstract}
\section{Introduction}
\label{sec:intro}
\subsection{Motivation}
\label{subsec:motivation}
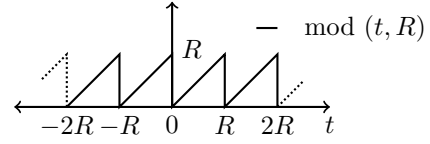
\begin{figure}[!t]
	\begin{center}
		\begin{tikzpicture}[scale=0.7]
			\draw[<->,thick] (-3,0)--(3,0) node[anchor=north]{$t$};
			\draw (0,0) node[anchor=north]{$0$};
			\draw (0,1.1) node[anchor=west] {$R$};
			\draw (1,0) node[anchor=north]{$R$};
			\draw (2,0) node[anchor=north] {$2R$};
			\draw (-1,0) node[anchor=north]{$-R$};
			\draw (-2,0) node[anchor=north] {$-2R$};
			\draw[->,thick] (0,0)--(0,2);
			\draw[] (2,1.5) node[anchor=west] {{$\mod(t,R)$}};
			\draw[thick] (1.6,1.5) -- (2,1.5);
			\draw[thick] (-2,0) --(-1,1)-| (-1,0) -- (0,1) -| (0,0) --(1,1)-| (1,0) -- (2,1) -| (2,0);
			\draw[densely dotted,thick] (2,0)--(2.5,0.5);
			\draw[densely dotted,thick] (-2,0)|-(-2,1) -- (-2.5,0.5);
			\end{tikzpicture}
	\end{center}
	\caption{\emph{The modulo transfer function.}}
	\label{fig:orgmodop}
\end{figure}

The problem of reconstructing a signal (or image) from (possibly) nonlinear observations is widely encountered in standard signal acquisition and imaging systems. Our focus in this paper is the problem of signal reconstruction from \textit{modulo} measurements, where the modulo operation with respect to a positive real valued parameter $R$ returns the (fractional) remainder after division by $R$. See Fig.~\ref{fig:orgmodop} for an illustration.


Formally, we consider a high dimensional signal (or image) $\mb{x}^* \in \R^n$. We are given modulo measurements of $\mb{x^*}$, that is, for each measurement vector $\mb{a_i} \in \R^n$, we observe:
\begin{equation}
y_i=\mod(\langle \mathbf{a_i} \cdot \mathbf{x^*} \rangle,R)~~i = \{1,2,...,m\}, 
\label{eq:modmeas0}
\end{equation} 
The task is to recover $\mb{x^*}$ using the modulo measurements $\mb{y}$ and knowledge the measurement matrix $\mathbf{A} = \left[\mathbf{a_1~a_2~...~a_m}\right]^\t$. 



This specific form of signal recovery is gaining rapid interest in recent times. Recently, the use of a novel imaging sensor that wraps the data in a periodical manner  has been shown to overcome certain hardware limitations of typical imaging systems \cite{Bhandari,ICCP15_Zhao,Shah,Cucuringu2017}. Many image acquisition systems suffer from the problem of limited dynamic range; however, real-world signals can contain a large range of intensity levels, and if tuned incorrectly, most intensity levels can lie in the saturation region of the sensors, causing loss of information through signal clipping. The problem gets amplified in the case of multiplexed linear imaging systems (such as compressive cameras or coded aperture systems), where required dynamic range is very high because of the fact that each linear measurement is a weighted aggregation of the original image intensity values. 

The standard solution to this issue is to improve sensor dynamic range via enhanced hardware; this, of course, can be expensive. An intriguing alternative is to deploy special digital \emph{modulo} sensors~\cite{rheejoo,kavusi2004quantitative,sasagawa2016implantable,yamaguchi2016implantable}. As the name suggests, such a sensor wraps each signal measurement around a scalar parameter $R$ that reflects the dynamic range. However, this also makes the forward model \eqref{eq:modmeas0} highly nonlinear and the reconstruction problem highly ill-posed. The approach of~\cite{Bhandari,ICCP15_Zhao} resolves this problem by assuming \emph{overcomplete} observations, meaning that the number of measurements $m$ is higher than the ambient dimension $n$ of the signal itself. For the cases where $m$ and $n$ are large, this requirement puts a heavy burden on computation and storage. 

In contrast, our focus is on solving the the inverse problem~\eqref{eq:modmeas0} with very few number of samples, {i.e.}, we are interested in the case $m \ll n$. While this makes the problem even more ill-posed, we show that such a barrier can be avoided if we assume that the underlying signal obeys a certain low-dimensional structure. In this paper, we focus on the \emph{sparsity} assumption on the underlying signal, but our techniques could be extended to other signal structures. Further, for simplicity, we assume that our forward model is limited to only two modulo periods, as shown in the Fig.~\ref{fig:compare}(a). Such a simplified variation of the modulo function already inherits much of the challenging aspects of the original recovery problem. Intuitively, this simplification requires that the value of dynamic range parameter $R$ should be large enough so that all the measurements $\langle \mathbf{a_i} \cdot \mathbf{x^*} \rangle$ can be covered within the domain of operation of the modulo function, \textit{i.e.}, $\langle \mathbf{a_i} \cdot \mathbf{x^*} \rangle \in [-R,R]~\forall i \in \{1,2,..,m\}$.

\subsection{Our contributions}
In this paper, we propose a recovery algorithm for exact reconstruction of signals from modulo measurements of the form \eqref{eq:modmeas0}. We refer our algorithm as \emph{MoRAM}, short for \emph{Modulo Recovery using Alternating Minimization}. The key idea in our approach is to identify and draw parallels between modulo recovery and the problem of \emph{phase retrieval}. Indeed, this connection enables us to bring in algorithmic ideas from classical phase retrieval, which also helps in our analysis. 

\begin{figure}[h]
	\begin{center}
		\begin{tabular}{cc}
			\begin{tikzpicture}[scale=0.45, every node/.style={scale=0.7}]
			\draw[<->] (-4,0) -- (4,0) node[right] {$t$};
			\draw[->] (0,-1) -- (0,4);
			\draw[scale=0.5, dashed, blue, thick] (4,4)--(4,0) node[above right]{$R$};
			\draw (1.8,-0.5) node(below) {$\sgn(t) = -1$};
			\draw (-1.8,-0.5) node(below) {$\sgn(t) = 1$};
			\draw[scale=0.5,blue,thick] (1.5,6.5)--(2.5,6.5) node[right,black]{$f(t)$};
			\draw (0,-1.5) node(right) {$f(t) = \mod(t,R)$};
			\draw[scale=0.5,domain=-4:0,smooth,variable=\x,blue, thick] plot ({\x},{\x+4});
			\draw[scale=0.5,domain=0:4,smooth,variable=\x,blue, thick]  plot ({\x},{\x});
			\end{tikzpicture} &
			
			\begin{tikzpicture}[scale=0.45, every node/.style={scale=0.7}]
			\draw[<->] (-4,0) -- (4,0) node[right] {$t$};
			\draw[->] (0,-1) -- (0,4);
			\draw (1.8,-0.5) node(below) {$\sgn(t) = -1$};
			\draw (-1.8,-0.5) node(below) {$\sgn(t) = 1$};
			\draw[scale=0.5,red,thick] (1.5,6.5)--(2.5,6.5) node[right,black]{$g(t)$};
			\draw (0,-1.5) node(right) {$g(t)=\mathrm{abs}(t)$};
			\draw[scale=0.5,domain=0:4,smooth,variable=\x,red,thick]  plot ({\x},{\x});
			\draw[scale=0.5,domain=-4:0,smooth,variable=\x,red, thick]  plot ({\x},{-\x});
			\end{tikzpicture} \\
			(a) & (b)
		\end{tabular}
	\end{center}
	\caption{\emph{Comparison between (a) modulo function ($f(t) = \mod(t,R)$); and (b) absolute value function ($g(t) = \mathrm{abs}(t)$).}}
	\label{fig:compare}
\end{figure}
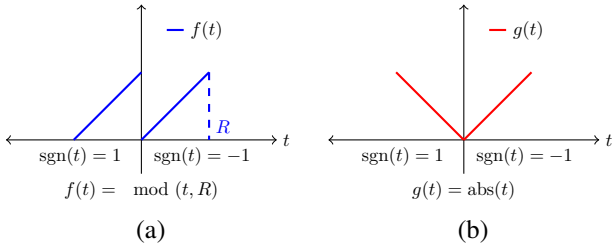

Phase retrieval has its roots in several classical imaging problems, but has attracted renewed interest of late. There, we are given observations of the form:
\[
y_i= | \langle \mathbf{a_i} , \mathbf{x^*} \rangle|,~~i = 1, 2, \ldots, m,
\]
and are tasked with reconstructing $\mathbf{x^*}$.  While these two different class of problems appear different at face value, the common theme is the need of undoing the effect of a piecewise linear transfer function applied to the observations. See Fig.~\ref{fig:compare} for a comparison.
Both the functions are identical to the identity function in the positive half, but differ significantly in the negative half. Solving the phase retrieval problem is essentially equivalent to retrieving the phase ($\text{sign}\left(y_i\right)$) corresponding to each measurement $y_i$. However, the phase can take only two values: $1$ if $t \geq 0$, or $-1$ if $t < 0$. Along the same lines, for modulo recovery case, the challenge is to identify the bin-index for each measurement. 
Estimating the bin-index correctly lets us ``unravel'' the modulo transfer function, thereby enabling signal recovery.

At the same time, several essential differences between the two problems restrict us from using phase retrieval algorithms as-is for the modulo reconstruction problem. The absolute value function can be represented as a \emph{multiplicative} transfer function (with the multiplying factors being the signs of the linear measurements), while the modulo function adds a constant value ($R$) to negative inputs. 
Therefore, the estimation procedures propagate very differently in the two cases. In the case of phase retrieval, a wrongly estimated phase induces an error that increases \emph{linearly} with the magnitude of each measurement. 
On the other hand, for modulo recovery problem, the error induced by an incorrect bin-index is $R$ (or larger), irrespective of the measurement. Therefore, existing algorithms for phase retrieval perform rather poorly for our problem (both in theory and practice). 

We resolve this issue by making non-trivial modifications to existing phase retrieval algorithms that better exploit the structure of modulo reconstruction. We also provide analytical proofs for recovering the underlying signal using our algorithm, and show that such a recovery can be performed using an (essentially) optimal number of observations, provided certain standard assumptions are met. To the best of our knowledge we are the first to pursue this type of approach for modulo recovery problems with \emph{generic} linear measurements, distinguishing us from previous work~\cite{ICCP15_Zhao,Bhandari}. 

\subsection{Techniques}

The basic approach in our proposed (MoRAM) algorithm is similar to several recent non-convex phase retrieval approaches. We pursue two stages. 

In the first stage, we identify a good initial estimated signal $\mb{x^0}$ that that lies (relatively) close to the true signal $\mb{x^*}$. A commonly used initialization technique for phase retrieval is \emph{spectral initialization} as described in \cite{netrapalli2013phase}. However, that does not seem to succeed in our case, due to markedly different behavior of the modulo transfer function. 
Instead, we introduce a novel approach of measurement \emph{correction} by comparing our observed measurements with typical density plots of Gaussian observations. Given access to such corrected measurements, $\mb{x^0}$ can be calculated simply by using a first-order estimator. This method is intuitive, yet provides 
	a provable guarantee for getting an initial vector that is close to the true signal. 

In the second stage, we refine this coarse initial estimate to recover the true underlying signal. Again, we follow an alternating-minimization (AltMin) approach inspired from phase retrieval algorithms~(such as \cite{netrapalli2013phase}) that estimates the signal and the measurement bin-indices alternatively. However, as mentioned above, any estimation errors incurred in the first step induces fairly large additive errors (proportional to the dynamic range parameter $R$.) We resolve this issue by using a \emph{robust} form of alternating-minimization (specifically, the Justice Pursuit algorithm~\cite{Laska2009}). We prove that AltMin, based on Justice Pursuit, succeeds provided the number of wrongly estimated bin-indices in the beginning is a small fraction of the total number of measurements. This gives us a natural radius for initialization, and also leads to provable sample-complexity upper bounds.   

\subsection{Paper organization} 
The reminder of this paper is organized as follows. In Section~\ref{sec:prior}, we briefly discuss the prior work. Section~\ref{sec:prelim} contains notation and mathematical model used for our analysis. In Section~\ref{sec:algo}, we introduce the MoRAM algorithm and provide a theoretical analysis of its performance. We demonstrate the performance of our algorithm by providing series of numerical experiments in Section~\ref{sec:exp}. Section~\ref{sec:disc} provides concluding remarks.
\section{Prior work}
\label{sec:prior}

We now provide a brief overview of related prior work. At a high level, our algorithmic development follows two (hitherto disconnected) streams of work in the signal processing literature.

\subsubsection*{Phase retrieval} As stated earlier, in this paper we borrow algorithmic ideas from previously proposed solutions for phase retrieval to solve the modulo recovery problem. Being a classical problem with a variety of applications, phase retrieval has been studied significantly in past few years. Approaches to solve this problem can be broadly classified into two categories: convex and non-convex. 

Convex approaches usually consist of solving a constrained optimization problem after lifting the true signal $\mb{x^*}$ in higher dimensional space. The seminal PhaseLift formulation \cite{candes2013phaselift} and its variations \cite{gross2017improved}, \cite{candes2015phasediff} come under this category. Recently,~\cite{Bahmani2016PhaseRM, Goldstein2018PhaseMaxCP} proposed a convex optimization based algorithm for solving the phase retrieval problem that doesn't lift the underlying signal and hence does not square the number of variables. The precise analysis of it has been derived in~\cite{Dhifallah2017PhaseRV,Salehi2018APA}. Typical non-convex approaches involve finding a good initialization, followed by iterative minimization of a loss function. Approaches based on Wirtinger Flow \cite{candes2015phase, zhang2016reshaped,  chen2015solving, cai2016optimal} and Amplitude flow \cite{wang2016sparse,wang2016solving} come under this category. 

In recent works, extending phase retrieval algorithms to situations where the underlying signal exhibits a sparse representation in some known basis has attracted interest. Convex approaches for sparse phase retrieval include \cite{ohlsson2012cprl, li2013sparse,bahmani2015efficient,jaganathan2012recovery}. Similarly, non-convex approaches for sparse phase retrieval include \cite{netrapalli2013phase, cai2016optimal, wang2016sparse}. Our approach in this paper towards solving the modulo recovery problem can be viewed as a complement to the non-convex sparse phase retrieval framework advocated in \cite{Jagatap2017}. 

\subsubsection*{Modulo recovery} The modulo recovery problem is also known in the classical signal processing literature
as phase unwrapping. The algorithm proposed in \cite{bioucas2007phase} is specialized to images, and employs graph cuts for phase unwrapping from a single modulo measurement per pixel. However, the inherent assumption there is that the input image has very few sharp discontinuities, and this makes it unsuitable for practical situations with textured images. Our work is motivated by the recent work of \cite{ICCP15_Zhao} on high dynamic range (HDR) imaging using a modulo camera sensor. For image reconstruction using multiple measurements, they propose the multi-shot UHDR recovery algorithm, with follow-ups developed further in \cite{Lang2017}. However, the multi-shot approach depends on carefully designed camera exposures, while our approach succeeds for non-designed (generic) linear observations; moreover, they do not include sparsity in their model reconstructions. In our previous work \cite{Shah}, we proposed a different extension based on \cite{ICCP15_Zhao, soltani2017stable} for signal recovery from quantized modulo measurements, which can also be adapted for sparse measurements, but there too the measurements need to be carefully designed.

In the literature, several authors have attempted to theoretically understand the modulo recovery problem. Given modulo-transformed time-domain samples of a band-limited function, \cite{Bhandari} provides a stable algorithm for perfect recovery of the signal and also proves sufficiency conditions that guarantees the perfect recovery. \cite{Cucuringu2017} formulates and solves an QCQP problem with non-convex constraints for denoising the modulo-1 samples of the unknown function along with providing a least-square based modulo recovery algorithm.. However, both these methods relay on the smoothness of the band-limited function as a prior structure on the signal, and as such it is unclear how to extend their use to more complex modeling priors (such as sparsity in a given basis).

In recent works,~\cite{Bhandari2018UnlimitedSO} proposed unlimited sampling algorithm for sparse signals. Similar to~\cite{Bhandari}, it also exploits the bandlimitedness by considering the low-pass filtered version of the sparse signal, and thus differs from our random measurements setup. In~\cite{Musa2018GeneralizedAM}, modulo recovery from Gaussian random measurements is considered, however, it assumes the true signal to be distributed as mixed Bernoulli-Gaussian distribution, which is impractical in real world imaging scenarios.

For a qualitative comparison of our MoRAM method with existing approaches, refer Table~\ref{tab:compare}. The table suggests that the previous approaches varied from the Nyquist-Shannon sampling setup only along the amplitude dimension, as they rely on band-limitedness of the signal and uniform sampling grid. We vary the sampling setup along both the amplitude and time dimensions by incorporating sparsity in our model, which enables us to work with non-uniform sampling grid (random measurements) and achieve a provable sub-Nyquist sample complexity.

\renewcommand{\arraystretch}{0.1}
\begin{table*}[t]
	\centering
	\caption{Comparison of MoRAM with existing modulo recovery methods. \label{tab:comp}}
	\begin{tabular}{p{4.7cm}>{\centering\arraybackslash}p{3cm}>{\centering\arraybackslash}p{2.05cm}>{\centering\arraybackslash}p{2.7cm}>{\centering\arraybackslash}p{3.35cm}}
		\toprule
		~& Unlimited Sampling~\cite{Bhandari}& OLS Method~\cite{Cucuringu2017} & multishot UHDR~\cite{ICCP15_Zhao}  & \textbf{MoRAM (our approach)} \\ \cmidrule(lr){1-5}
        \cmidrule(lr){1-5}
		Assumption on structure of signal  & Bandlimited   & Bandlimited &  No assumptions & Sparsity     \\ \cmidrule(lr){1-5}
		Sampling scheme  & uniform grid &  uniform grid  & (carefully chosen) linear measurements   & 
		random linear measurements \\ \cmidrule(lr){1-5}
		Sample complexity & oversampled, $\mathcal{O}(n)$    & --    & oversampled, $\mathcal{O}(n)$   & undersampled,~$\mathcal{O}(s\log(n))$ \\ \cmidrule(lr){1-5}
		Provides sample complexity bounds?  & Yes  & --  & No & Yes \\ \cmidrule(lr){1-5}
		Leverages Sparsity?  & No & No   & No    & Yes \\ \cmidrule(lr){1-5}
		(Theoretical) bound on dynamic range  & Unbounded   & Unbounded     & Unbounded   & $2R$  \\ 
		\bottomrule
	\end{tabular}	
\label{tab:compare}
\end{table*}

\section{Preliminaries}
\label{sec:prelim}
\subsection{Notation}
\label{subsec:nota}
Let us introduce some notation. We denote matrices using bold capital-case letters ($\mb{A,B}$), column vectors using bold-small case letters ($\mb{x,y,z}$ etc.) and scalars using non-bold letters ($R,m$ etc.). We use letters $C$ and $c$ to represent constants that are large enough and small enough respectively. We use $\mb{x}^\t,\mb{A}^\t$ to denote the transpose of the vector $\mb{x}$ and matrix $\mb{A}$ respectively. The cardinality of set $S$ is denoted by $\card(S)$.
We define the signum function as $\sgn(x) := \frac{x}{|x|}$ for every $x \in \R, x \neq 0$, with the convention that $\sgn(0)=1$. The $i^{th}$ element of the vector $\mb{x} \in \R^n$ is denoted by $x_{i}$. Similarly, $i^{th}$ row of the matrix $\mb{A} \in \R^{m \times n}$ is denoted by $\mb{a_i}$, while the element of $\mb{A}$ in the $i^{th}$ row and $j^{th}$ column is denoted as $a_{ij}$. The projection of $\mb{x} \in \R^n$ onto a set of coordinates $S$ is represented as $\mb{x}_S \in \R^n$, i.e., $\mb{x}_{S_j} = \mb{x}_j$ for $j \in S$, and $0$ elsewhere. 


\subsection{Mathematical model}
\label{subsec:model}
As depicted in Fig.~\ref{fig:compare}(a), we consider the modulo operation within 2 periods (one in the positive half and one in the negative half). We assume  that the value of dynamic range parameter $R$ is large enough so that all the measurements $\langle \mathbf{a_i} \cdot \mathbf{x^*} \rangle$ are covered within the domain of operation of modulo function. Rewriting in terms of the signum function, the (variation of) modulo function under consideration can be defined as: 
$$
f(t) := t+\left( \frac{1-\sgn(t)}{2}\right)R.
$$
One can easily notice that the modulo operation in this case is nothing but an addition of scalar $R$ if the input is negative, while the non-negative inputs remain unaffected by it. If we divide the number line in these two bins, then the coefficient of $R$ in above equation can be seen as a bin-index, a binary variable which takes value $0$ when $\sgn(t)=1$, or $1$ when $\sgn(t)=-1$.
Inserting the definition of $f$ in the measurement model of Eq.~\ref{eq:modmeas0} gives,
\begin{equation}
y_i= \langle \mathbf{a_i} \cdot \mathbf{x^*} \rangle+\left( \frac{1-\sgn(\langle \mathbf{a_i} \cdot \mathbf{x^*} \rangle)}{2}\right)R,~~i = \{1,..,m\}.
\label{eq:modmeas2}
\end{equation} 
We can rewrite Eq.~\ref{eq:modmeas2} using a bin-index vector $\mb{p} \in \{0,1\}^m$. Each element of the true bin-index vector $\mb{p}^*$ is given as,
$$
p^*_i = \frac{1-\sgn(\langle \mathbf{a_i} \cdot \mathbf{x^*} \rangle)}{2},~~i = \{1,..,m\}.
$$

If we ignore the presence of modulo operation in above formulation, then it reduces to a standard compressive sensing reconstruction problem. In that case, the compressed measurements $y_{c_i}$ would just be equal to $\langle \mathbf{a_i} \cdot \mathbf{x^*} \rangle$.    

While we have access only to the compressed modulo measurements $\mb{y}$, it is useful to write $\mb{y}$ in terms of true compressed measurements $\mb{y}_c$. Thus,
$$
y_i = \langle \mathbf{a_i} \cdot \mathbf{x^*} \rangle + p^*_iR = y_{c_i}+p^*_iR.
$$

It is evident that if we can recover $\mathbf{p^*}$ successfully, we can calculate the true compressed measurements $\langle \mathbf{a_i} \cdot \mathbf{x^*} \rangle$ and use them to reconstruct $\mathbf{x^*}$ with any sparse recovery algorithm such as CoSaMP~\cite{needell2010cosamp} or basis-pursuit~\cite{chen2001atomic,spgl1:2007,BergFriedlander:2008}.

\section{Sparse signal recovery}
\label{sec:algo}
Of course, the major challenge is that we do \emph{not} know the bin-index vector. In this section, we describe our algorithm to recover both $\mathbf{x^*}$ and $\mathbf{p^*}$, given $\mathbf{y, A, s, R}$.  Our algorithm \emph{MoRAM (Modulo Reconstruction with Alternating Minimization)} comprises of two stages: (i) an initialization stage, and (ii) descent stage via alternating minimization.

\subsection{Initialization by re-calculating the measurements}
\label{sec:init}

Similar to other non-convex approaches, MoRAM also requires an initial estimate $\mathbf{{x}^0}$ that is close to the true signal $\mathbf{{x}^*}$. We have several initialization techniques available; in phase retrieval, techniques such as spectral initialization are often used. However, the nature of the problem in our case is fundamentally different due to the non-linear \emph{additive} behavior of the modulo transfer function. To overcome this issue, we propose a method to re-calculate the true Gaussian measurements ($\mb{y_c}= \mb{Ax^*}$) from the available modulo measurements. 

The high level idea is to undo the nonlinear effect of modulo operation in a significant fraction of the total available measurements. To understand the method for such re-calculation, we will first try to understand the effect of modulo operation on the linear measurements.

\subsubsection{Effect of the modulo transfer function} 
\label{sec:modeff}
\begin{figure}[!t]
	\begin{center}
		\begin{tikzpicture}[scale=0.9, every node/.style={scale=0.9, font=\normalsize}]
		\def\normaltwo{\x,{3*1/exp(((\x)^2)/2)}}
		\def\normalone{\x,{3*1/exp(((\x-3.5)^2)/2)}}
		\def\normalsum{\x,{3*1/exp(((\x-3.5)^2)/2)+3*1/exp(((\x)^2)/2)}}
		\def\y{2}
		\def\fy{3*1/exp(((\y-3.5)^2)/2)}
		\fill [fill=gray!30] (1,0) -- plot[domain=1:3] (\normalsum) -- (3,0) -- cycle;
		\fill [fill=orange!60, opacity=1] (0,0) -- plot[domain=0:3] (\normaltwo) -- (3,0) -- cycle;
		\fill [fill=green!60, opacity=1] (1,0) -- plot[domain=0.5:3.5] (\normalone) -- (3.5,0) -- cycle;
		\fill [fill={rgb:orange!60,1;green!60,0.6}, opacity=1] (1,0) -- plot[domain=0.5:1.75] (\normalone) -- (1.75,0) -- cycle;
		\fill [fill={rgb:orange!60,1;green!60,0.6}, opacity=1] (1.75,0) -- plot[domain=1.75:3] (\normaltwo) -- (3,0) -- cycle;
		\draw[color=blue,domain=-0:3,dashed,thick] plot (\normaltwo) node[right] {};
		\draw[color=blue,domain=0.5:3.5,dashed,thick] plot (\normalone) node[right] {};
		\draw[color=blue,domain=-0:3.5,thick] plot (\normalsum) node[right] {};
		\draw[<->] (-0.5,0) -- (4.5,0);
		\draw (4.25,0) node[above] {$\mathbf{y}$};
		\draw[<->] (0,-0.8) -- (0,4);
		\draw[dashed] ({3.5},{3}) -- ({3.5},0);
		\draw[dashed] ({1.75},{3.5}) -- ({1.75},0);
		\draw (3.5,-0.5) node(below) {$R$};
		\draw (1.75,-0.5) node(below) {$R/2$};
		
		 \draw[<->](0,3.2)--(1.75,3.2);
		 \draw[<->](1.75,3.2)--(3.5,3.2);
		 \draw (0.85,3.5) node(above) {{$p^{init}_i=0$}};
		 \draw (2.65,3.5) node(above) {{$p^{init}_i=1$}};
	    \draw (1.75,0.15) -- (1.75,-0.15);
        \draw (3.50,0.15) -- (3.50,-0.15);
        \draw [dashed](3.5,3.15) -- (3.5,3.45);
		\end{tikzpicture}
	\end{center}
	\caption{\emph{Density plot of $\mod(\mathbf{Ax^*})$}. Best viewed in color.}
	\label{fig:hist2}
\end{figure}
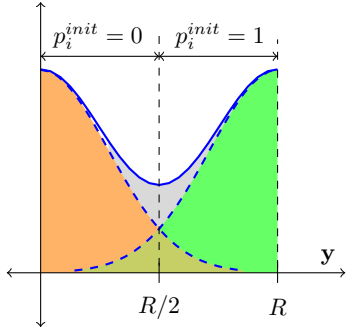

To provide some intuition, let us first examine the distribution of the $\mathbf{Ax^*}$(Fig.~\ref{fig:hist1}) and $\mathbf{\mod(\mathbf{Ax^*})}$ (shown in Fig.~\ref{fig:hist2}) to understand what information can be obtained from the modulo measurements. We are particularly interested in the case where the elements of $\mathbf{Ax^*}$ are small compared to the modulo range parameter $R$. 

 Note that the compressed measurements $\mathbf{y_c}$ follow the standard normal distribution, as $\mb{A}$ is Gaussian random matrix. These plots essentially depict the distribution of our observations \emph{before} and \emph{after} the modulo operation.

With reference to Fig.~\ref{fig:hist1}, we divide the compressed observations $\mb{y_c}$ in two sets: $\mb{y_{c,+}}$ contains all the non-negative observations (orange) with bin-index$=0$, while $\mb{y_{c,-}}$ contains all the negative ones (green) with bin-index$=1$.

As shown in Fig.~\ref{fig:hist2}, after modulo operation, the set $\mb{y}_{c,-}$ (green) shifts to the right by $R$ and gets concentrated in the right half ($[R/2,R]$); while the set $\mb{y}_{c,+}$ (orange) remains unaffected and concentrated in the left half ($[0,R/2]$). Thus, for some of the modulo measurements, their correct bin-index can be identified just by observing their magnitudes relative to the midpoint $R/2$. This leads us to obtain following maximum likelihood estimator for bin-indices ($\mb{p}$):

\begin{equation}
{p}^{init}_{i} = 
\begin{cases}
0,& \text{if } 0\leq y_i < R/2 \\
1,& \text{if } R/2 \leq y_i \leq R
\end{cases}
\label{eq:rcm}
\end{equation}

The $\mb{p^{init}}$ obtained with above method contains the correct values of bin-indices for many of the measurements, except for the ones concentrated within the ambiguous region in the center.

\begin{figure}[!t]
	\begin{center}
		\begin{tikzpicture}[scale=0.9, every node/.style={scale=0.9,font=\normalsize}]
			\def\normaltwo{\x,{3*1/exp(((\x)^2)/2)}}
			\def\y{4.4}
			
			\draw[color=blue,domain=-3.75:3.75,thick, samples=100] plot (\normaltwo) node[right] {};
			\draw[<->] (-3.75,0) -- (3.75,0);
			\draw (3.75,0.1) node[above left] {$\mathbf{y_c}$};
			\draw[<->] (0,-0.8) -- (0,4);
			\fill [fill=green!60] (0,0) -- plot[domain=-3.75:0] (\normaltwo) -- (0,0) -- cycle;
			\fill [fill=orange!60] (0,0) -- plot[domain=0:3.75] (\normaltwo) -- (0,0) -- cycle;
			\draw (-3.50,-0.5) node(below) {$-R$};
			\draw (3.50,-0.5) node(below) {$R$};
				
        \draw (-3.50,0.15) -- (-3.50,-0.15);
        \draw (3.50,0.15) -- (3.50,-0.15);
			\end{tikzpicture}
	\end{center}
	\caption{\emph{Density plot of $\mathbf{Ax^*}$}}
	\label{fig:hist1}
\end{figure}
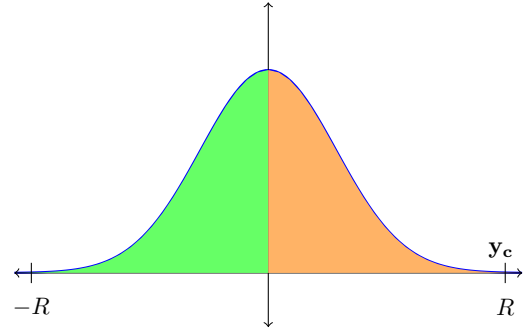

\begin{algorithm}[t]
	\caption{\textsc{MoRAM-initialization}}
	\label{alg:RCM}
	\begin{algorithmic}
		\State\textbf{Inputs:} $\mathbf{y}$, $\mathbf{A}$, $s$, $R$
		\State\textbf{Output:}  $\mb{x^0}$
		\For {$i= 0:m$}
%
		
		\State Calculate $p^{init}_i$ according to Eq.~\ref{eq:rcm}.
		\EndFor
		\State Calculate $\mb{y^{init}_c}$ according to Eq.~\ref{eq:y_init}.
		\State $\mb{x^0} \leftarrow H_s\left( \frac{1}{N}\sum_{i=1}^{N}y^{init}_{c,i}a_{i}\right)$
	\end{algorithmic}
\end{algorithm}

Once we identify the initial values of bin-index for the modulo measurements, we can calculate corrected measurements as,
\begin{align}
\mathbf{y^{init}_{c} = y + p^{init}R}.
\label{eq:y_init}
\end{align}
We use these corrected measurements $\mathbf{y^{init}_{c}}$ to calculate the initial estimate $\mathbf{{x}^0}$ with first order unbiased estimator.
\begin{equation}
\mb{x^0} = H_s\left( \frac{1}{N}\sum_{i=1}^{N}y^{init}_{c,i}a_{i}\right),
\label{eq:init}
\end{equation}
where $H_s$ denotes the hard thresholding operator that keeps the $s$ largest absolute entries of a vector and sets the other entries to zero.

\subsection{Alternating Minimization}
\label{sec:altmin}
\begin{algorithm}[t]
	\caption{\textsc{MoRAM-descent}}
	\label{alg:MoRAM}
	\begin{algorithmic}
		\State\textbf{Inputs:} $\mathbf{y}$, $\mathbf{A}$, $s$, $R$
		\State\textbf{Output:}  $\mb{x^T}$
		\State $m,n \leftarrow \mathrm{size}(\mathbf{A})$ 
		\State \textbf{Initialization}
		\State $\mathbf{x^0} \leftarrow \textrm{MoRAM-initialization}(\mathbf{y, A})$ 
		\State \textbf{Alternating Minimization}
		\For {$t =0:T$}
		\State $\mathbf{{p}^{t}} \leftarrow \frac{\mathbf{1}-\sgn(\langle \mathbf{A} \cdot \mathbf{x^t} \rangle)}{2}$
		\State $\mathbf{y^t_c} \leftarrow \mathbf{y} - \mathbf{p^t}R$
		\State $\mathbf{{x}^{t+1}}\leftarrow \small{JP(\frac{1}{\sqrt{m}}\begin{bmatrix} \mathbf{A} & \mathbf{I} \end{bmatrix},\frac{1}{\sqrt{m}}\mathbf{y^t_c},[\mathbf{x^t~~p^t}]^\t)}$.
		\EndFor
	\end{algorithmic}
\end{algorithm}

Using Eq.~\ref{eq:init}, we calculate the initial estimate of the signal $\mathbf{{x}^0}$ which is relatively close to the true vector $\mathbf{x^*}$. Starting with $\mathbf{{x}^0}$, we calculate the estimates of $\mathbf{p}$ and $\mathbf{x}$ in an alternating fashion to converge to the original signal $\mathbf{x^*}$. At each iteration of alternating-minimization, we use the current estimate of the signal ${\mathbf{x^t}}$ to get the value of the bin-index vector $\mathbf{{p}^t}$ as following:
\begin{equation}
\mathbf{{p}^{t}} = \frac{\mathbf{1}-\sgn(\langle \mathbf{A} \cdot \mathbf{x^t} \rangle)}{2}.
\label{step1}
\end{equation}

Given that $\mathbf{x^0}$ is close to $\mathbf{x^*}$, we expect that $\mathbf{p^0}$ would also be close to $\mathbf{p^*}$. Ideally, we would calculate the correct compressed measurements $\mathbf{y^t_c}$ using $\mathbf{p^t}$, and use $\mathbf{y^t_c}$ with any popular compressive recovery algorithms such as CoSaMP or basis pursuit to calculate the next estimate $\mathbf{{x}^{t+1}}$. Thus,

$$
\mathbf{y^t_c} = \langle \mathbf{A}\mathbf{x^{t+1}} \rangle = \mathbf{y} - \mathbf{p^t}R,
$$

$$
\mathbf{{x}^{t+1}} = \argmin_{\mathbf{x} \in \mathcal{M}_s}\norm{\mathbf{Ax} - \mathbf{y^t_c}}_2^2, 
$$
where $\mathcal{M}_s$ denotes the set of $s$-sparse vectors in $\mathbb{R}^n$. Note that sparsity is only one of several signal models that can be used here, and in principle a rather similar formulation would extend to cases where $\mathcal{M}$ denotes any other structured sparsity model~\cite{modelcs}.

However, it should be noted that the ``bin'' error $\mathbf{d^t} = \mathbf{p^t - p^*}$, even if small, would significantly impact the correction step that constructs $\mathbf{y^t_c}$, as each incorrect bin-index would add a noise of the magnitude $R$ in $\mathbf{y^t_c}$. Our experiments suggest that the typical sparse recovery algorithms are not robust enough to cope up with such large errors in $\mathbf{y^t_c}$. To tackle this issue, we employ an outlier-robust sparse recovery method \cite{Laska2009}. We consider the fact that the nature of the error $\mathbf{d^t}$ is sparse with sparsity $s_{dt}=\norm{\mb{d^t}}_0$; and each erroneous element of $\mathbf{p}$ adds a noise of the magnitude $R$ in $\mathbf{y^t_c}$.

Rewriting in terms of Justice Pursuit, the recovery problem now becomes problem becomes,

$$
\mathbf{{x}^{t+1}}=\argmin_{[\mathbf{x~d}]^\t \in \mathcal{M}_{s+s_{dt}}}\norm{\begin{bmatrix} \mathbf{A} & \mathbf{I} \end{bmatrix} \begin{bmatrix} \mathbf{x} \\ \mathbf{d} \end{bmatrix} - \mathbf{y^t_c}}_2^2, 
$$

 However, the sparsity of $\mathbf{d^t}$ is unknown, suggesting that greedy sparse recovery methods cannot be directly used without an additional hyper-parameter. Thus, we employ basis pursuit~\cite{candes2006compressive} which does not rely on sparsity. The robust formulation of basis pursuit is referred as Justice Pursuit (JP) \cite{Laska2009}, specified in Eq.~\ref{eq:jp}.
\begin{equation}
\implies \mathbf{{x^{t+1}}} = JP(\frac{1}{\sqrt{m}}\begin{bmatrix} \mathbf{A} & \mathbf{I} \end{bmatrix},\frac{1}{\sqrt{m}}\mathbf{y^t_c},[\mathbf{x^t~~p^t}]^\t).
\label{eq:jp}
\end{equation}
Proceeding this way, we repeat the steps of bin-index calculation (as in Eq.~\ref{step1}) and sparse recovery (Eq.~\ref{eq:jp}) altenatingly for $\mathrm{T}$ iterations. Our algorithm is able to achieve convergence to the true underlying signal, as supported by the results in the experiments section.



%

\section{Mathematical Analysis}
\label{sec:mathanalysis}

Before presenting experimental validation of our proposed MoRAM algorithm, we now perform a theoretical analysis of the descent stage of our algorithm. We assume the availability of an initial estimate $\mathbf{x^0}$ that is close to $\mathbf{x^*}$, i.e. $\norm{\mb{x^0} - \mb{x^*}}_2 \leq \delta \norm{\mb{x^*}}_2$. In our case, our initialization step (in Alg. \ref{alg:MoRAM}) provide such $\mb{x^0}$.
We perform alternating-minimization as described in~\ref{alg:MoRAM}, starting with $\mb{x^0}$ calculated using Alg.~\ref{alg:RCM}. For simplicity, we limit our analysis of the convergence to only one AltMin iteration. In fact, according to our theoretical analysis, if initialized closely enough, one iteration of AltMin suffices for exact signal recovery with sufficiently many measurements. However, in practice we have observed that our algorithm requires more than one AltMin iterations. 

The first step is to obtain the initial guess of the bin-index vector (say $\mb{p^0}$) using $\mb{x^0}$.
$$
\mathbf{{p}^{0}} = \frac{\mathbf{1}-\sgn(\langle \mathbf{A} \cdot \mathbf{x^0} \rangle)}{2}.
$$
If we try to undo the effect of modulo operation by adding back $R$ for the affected measurements based on the bin-index vector $\mb{p^0}$, it would introduce an additive error equal to $R$ corresponding to each of the incorrect bin-indices in $\mb{p^0}$.
$$
\mathbf{y^0_c} = \langle \mathbf{A}\mathbf{x^{0}} \rangle = \mathbf{y} - \mathbf{p^0}R,
$$

We show the guaranteed recovery of the true signal as the corruption in the first set of corrected measurements $\mb{y_c^0}$ can be modeled as sparse vector with sparsity less than or equal to $\lambda m$, with $c$ being a fraction that can be explicitly bounded.

To prove this, we first introduce the concept of \emph{binary $\epsilon$-stable embedding} as proposed by~\cite{Jacques2013}. Let $\mathcal{B}^m$ be a Boolean cube defined as $\mathcal{B}^m := \{-1,1\}^m$ and let $S^{n-1}:=\{\mb{x}\in \R^n : \norm{\mb{x}}_2 = 1\}$ be the unit hyper-sphere of dimension $n$.

\definition[Binary $\epsilon$-Stable Embedding]
{A mapping $F: \R^n \rightarrow \mathcal{B}^m$ is a binary $\epsilon$-stable embedding (B$\epsilon$SE) of order $s$ for sparse vectors if:
	$$
	d_S(\mb{x,y}) - \epsilon \leq d_H(F\mb{(x)},F\mb{(y)}) \leq d_S(\mb{x,y}) + \epsilon;
	$$
	for all $\mb{x,y} \in S^{n-1}$ with $|supp(x) \cup supp(y)|\leq s$.
}

In our case, let us define the mapping $F: \R^n \rightarrow \mathcal{B}^m$ as:
$$
F(\mb{x}):= \sgn(\mb{Ax});
$$
with $\mb{A} \sim \mathcal{N}^{m \times n}(0,1)$. We obtain:

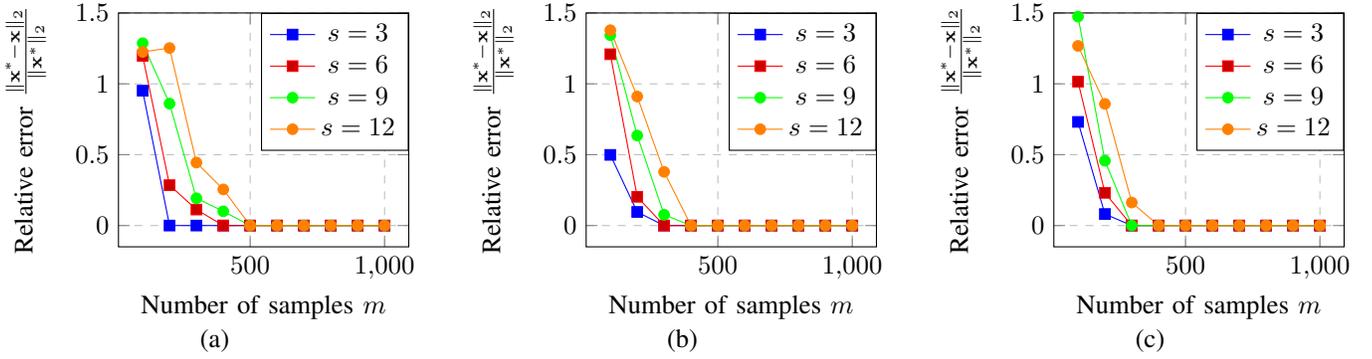
\begin{figure*}[h!]
	\begin{center}
		\begin{tabular}{ccc}
			\begin{tikzpicture}
\begin{axis}
[width=0.3\textwidth,
ymax = 1.5,
xlabel= Number of samples $m$, 
ylabel= Relative error $\frac{\norm{\mb{x^*-x}}_2}{\norm{\mb{x^*}}_2}$,
grid style = dashed,
grid=both,
legend style=
{
at={(1,1), 
anchor=top},
cells={align=center}, 
} 
]

\addplot[color=blue,mark=square*] plot coordinates {
(100,0.951634784301822)
(200,0.0000885116035094811)
(300,0.000081555309427471)
(400,0.0000848089765043806)
(500,0.0000728150352127786)
(600,0.0000513990857068759)
(700,0.0000504413020350128)
(800,0.0000530216699060703)
(900,0.0000700189218570823)
(1000,0.0000482831563501382)

};
\addlegendentry{$s=3$};

\addplot plot coordinates {
(100,1.19761376063554)
(200,0.285151912220181)
(300,0.112824660587099)
(400,0.0000900722262841476)
(500,0.0000834948390804258)
(600,0.0000822810668822694)
(700,0.0000674063715962388)
(800,0.0000575811497064516)
(900,0.0000682040611974146)
(1000,0.0000652156411921387)

};

\addlegendentry{$s=6$};

\addplot[color=green,mark=*] plot coordinates {
(100,1.28676059229197)
(200,0.860528910694914)
(300,0.192632429033674)
(400,0.100622445508152)
(500,0.0000797917821215397)
(600,0.0000868943512927315)
(700,0.0000831613767862367)
(800,0.0000691424927254167)
(900,0.0000697908622513554)
(1000,0.0000663731955980047)
};
\addlegendentry{$s=9$};

\addplot[color=orange,mark=*] plot coordinates {
(100,1.22525398257066)
(200,1.25271075983445)
(300,0.444364309906652)
(400,0.255141779884858)
(500,0.0000809007126847413)
(600,0.00010414145725093)
(700,0.0000923774185769297)
(800,0.0000915880164921741)
(900,0.0000776084169563498)
(1000,0.0000796310977264967)
};
\addlegendentry{$s=12$};


\end{axis}
\end{tikzpicture}
			&
			\begin{tikzpicture}
\begin{axis}
[width=0.3\textwidth,
ymax = 1.5,
xlabel= Number of samples $m$, 
ylabel= Relative error $\frac{\norm{\mb{x^*-x}}_2}{\norm{\mb{x^*}}_2}$,
grid style = dashed,
grid=both,
legend style=
{
at={(1,1), 
anchor=top},
cells={align=center}, 
} 
]

\addplot[color=blue,mark=square*] plot coordinates {
(100,0.498710161320585)
(200,0.0959538320002874)
(300,0.0000792295115649365)
(400,0.0000668741260877727)
(500,0.0000594711406822091)
(600,0.0000664390913303021)
(700,0.000059171865932364)
(800,0.0000536302147038852)
(900,0.0000487084123049625)
(1000,0.0000521328824625665)
};
\addlegendentry{$s=3$};

\addplot plot coordinates {
(100,1.2088704442973)
(200,0.202928802686383)
(300,0.0000890186779148784)
(400,0.0000910284011492659)
(500,0.0000923392498794239)
(600,0.0000791130903286572)
(700,0.0000675384947192539)
(800,0.0000661813455226514)
(900,0.0000671129023455458)
(1000,0.00006191654588485)
};

\addlegendentry{$s=6$};

\addplot[color=green,mark=*] plot coordinates {
(100,1.34347903915758)
(200,0.635274088925638)
(300,0.0766072949062092)
(400,0.000107519249510194)
(500,0.000102968285355943)
(600,0.0000810073754120081)
(700,0.0000745687663232692)
(800,0.000082397219376368)
(900,0.000068122133414809)
(1000,0.000066227771710853)
};
\addlegendentry{$s=9$};

\addplot[color=orange,mark=*] plot coordinates {
(100,1.37868747598793)
(200,0.910642080920763)
(300,0.379016640491203)
(400,0.000111800813242959)
(500,0.000094380953300525)
(600,0.0000778637044755454)
(700,0.0000998297574449876)
(800,0.0000882835763997106)
(900,0.0000765674476821649)
(1000,0.0000697311360773821)
};
\addlegendentry{$s=12$};


\end{axis}
\end{tikzpicture}
			&
			\begin{tikzpicture}
\begin{axis}
[width=0.30\textwidth,
ymax = 1.5,
xlabel= Number of samples $m$, 
ylabel= Relative error $\frac{\norm{\mb{x^*-x}}_2}{\norm{\mb{x^*}}_2}$,
grid style = dashed,
grid=both,
legend style=
{
	at={(1,1), 
		anchor=top},
	cells={align=center}, 
} 
]
\addplot[color=blue,mark=square*] plot coordinates {
(100,0.731398752407562)
(200,0.0813273619236995)
(300,0.0000767160765564396)
(400,0.0000727029062020334)
(500,0.000059814724782405)
(600,0.0000607377471274783)
(700,0.0000674078424157802)
(800,0.0000674180774975706)
(900,0.000063974897910302)
(1000,0.0000692817552090165)
};
\addlegendentry{$s=3$};

\addplot plot coordinates {
(100,1.01542896871115)
(200,0.230784211843143)
(300,0.000095988324114983)
(400,0.0000961055645259853)
(500,0.0000842880281224106)
(600,0.0000919952357507636)
(700,0.0000581074207742978)
(800,0.0000745427871300645)
(900,0.0000560512267283863)
(1000,0.0000571409494890391)
};

\addlegendentry{$s=6$};

\addplot[color=green,mark=*] plot coordinates {
(100,1.47457024298556)
(200,0.457072962810423)
(300,0.000100794494567937)
(400,0.000086050231662183)
(500,0.0000962448748572098)
(600,0.000083847748771777)
(700,0.000083460613651214)
(800,0.0000709897962825585)
(900,0.0000514897266332198)
(1000,0.0000696959287812024)

};
\addlegendentry{$s=9$};

\addplot[color=orange,mark=*] plot coordinates {
(100,1.2678278132646)
(200,0.858551024708902)
(300,0.162470095823513)
(400,0.0000895892548131344)
(500,0.0001101271180892)
(600,0.000100260074637336)
(700,0.0000973566189108632)
(800,0.0000819451823594468)
(900,0.0000889954295819448)
(1000,0.0000756339643144772)
};
\addlegendentry{$s=12$};


\end{axis}
\end{tikzpicture} \\
			(a) & (b) & (c) \\
		\end{tabular}
	\end{center}
	\caption{{Mean relative reconstruction error vs no. of measurements $(m)$ for MoRAM with $\norm{\mb{x^*}}_2=1,n=1000$, and (a) $R=4$; (b) $R=4.25$; (c) $R=4.5$.}}
	\label{fig:plot}
\end{figure*}

\begin{figure*}[h!]
	\begin{center}
		\setlength{\tabcolsep}{1pt}
		\begin{tabular}{ccccc}
			\rotatebox{90}{$~~~~~~~m=4000$} &
			\includegraphics[width=0.22\linewidth]{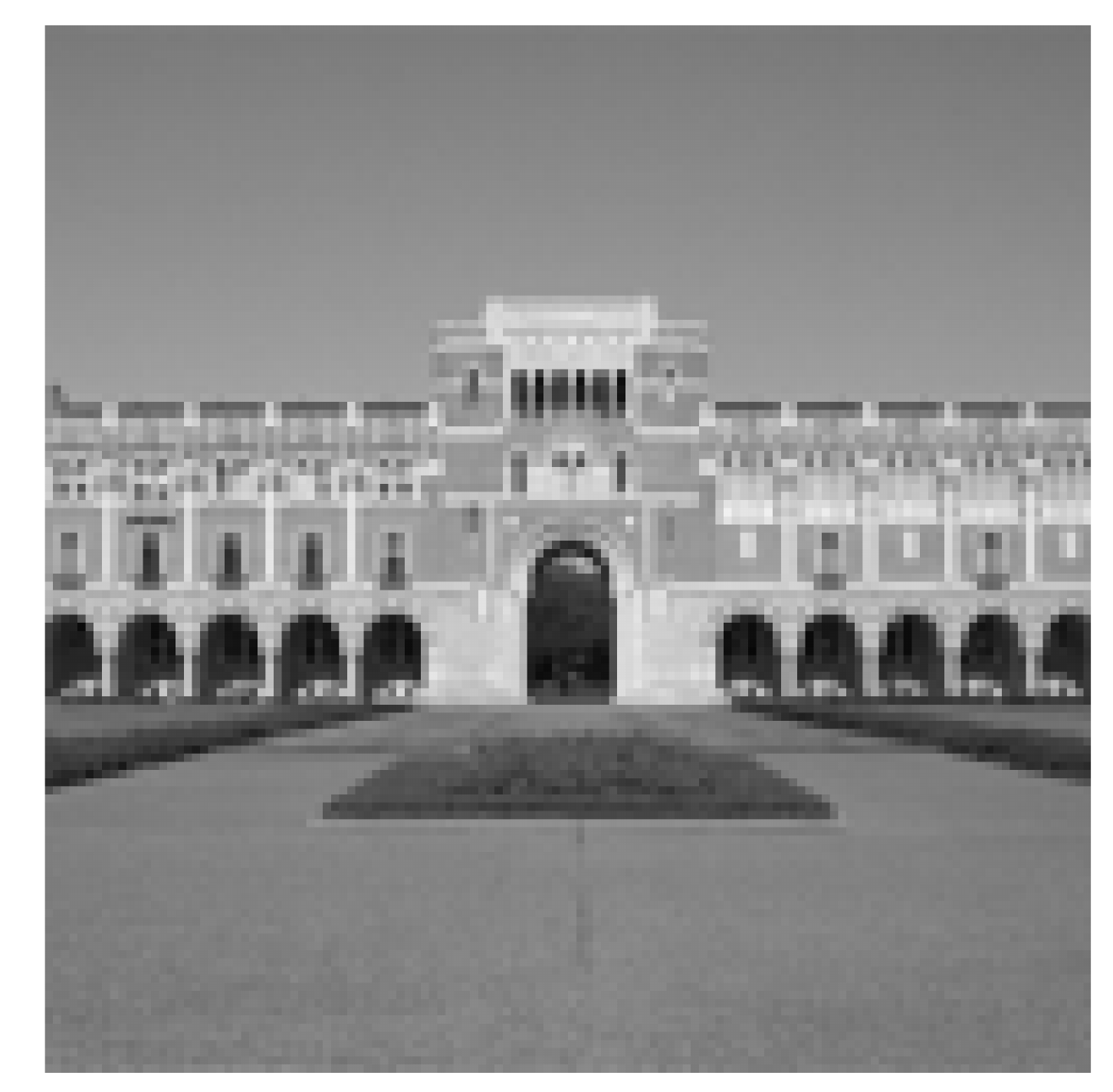} &
			\includegraphics[width=0.22\linewidth]{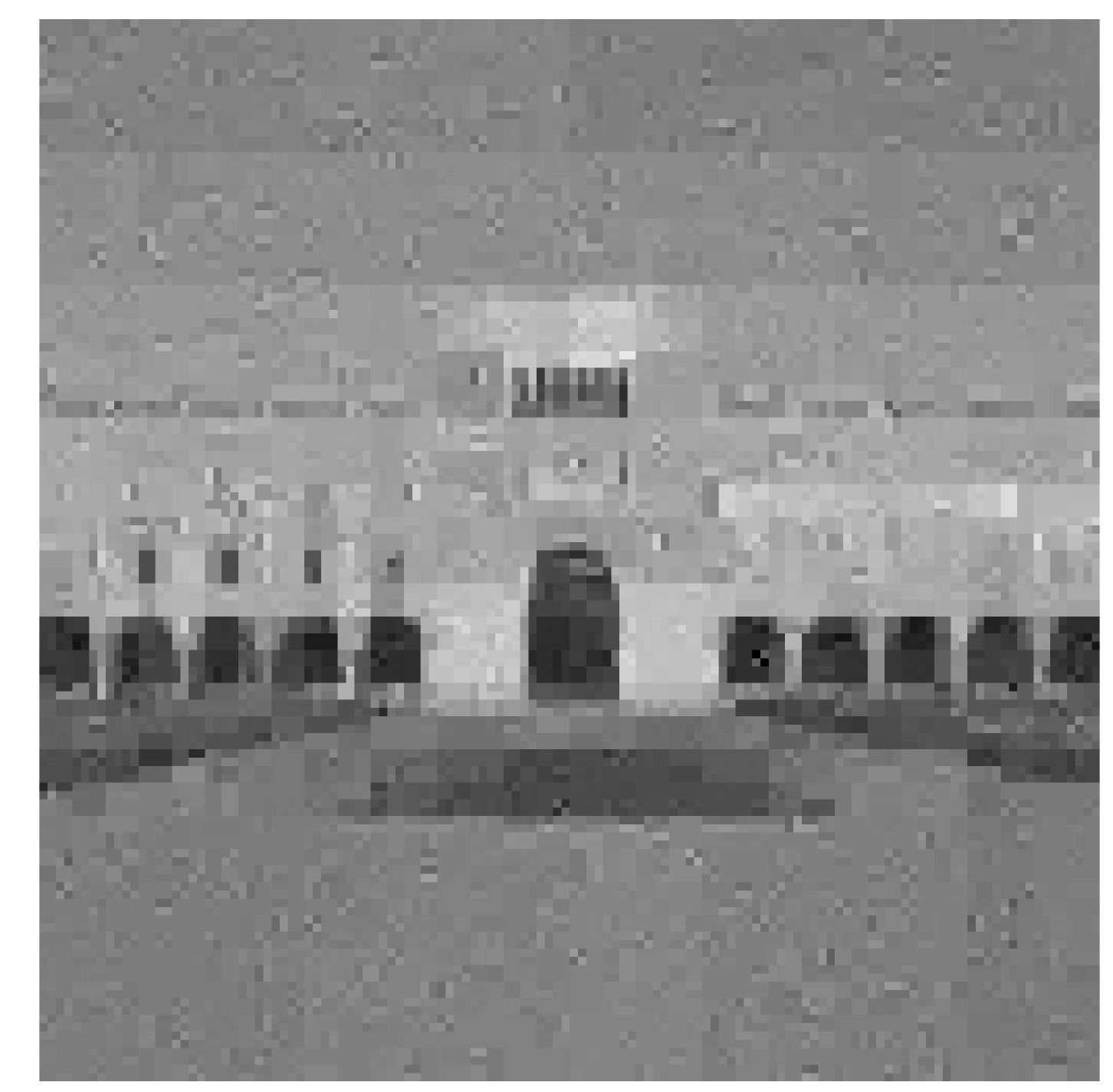} & 
			\includegraphics[width=0.22\linewidth]{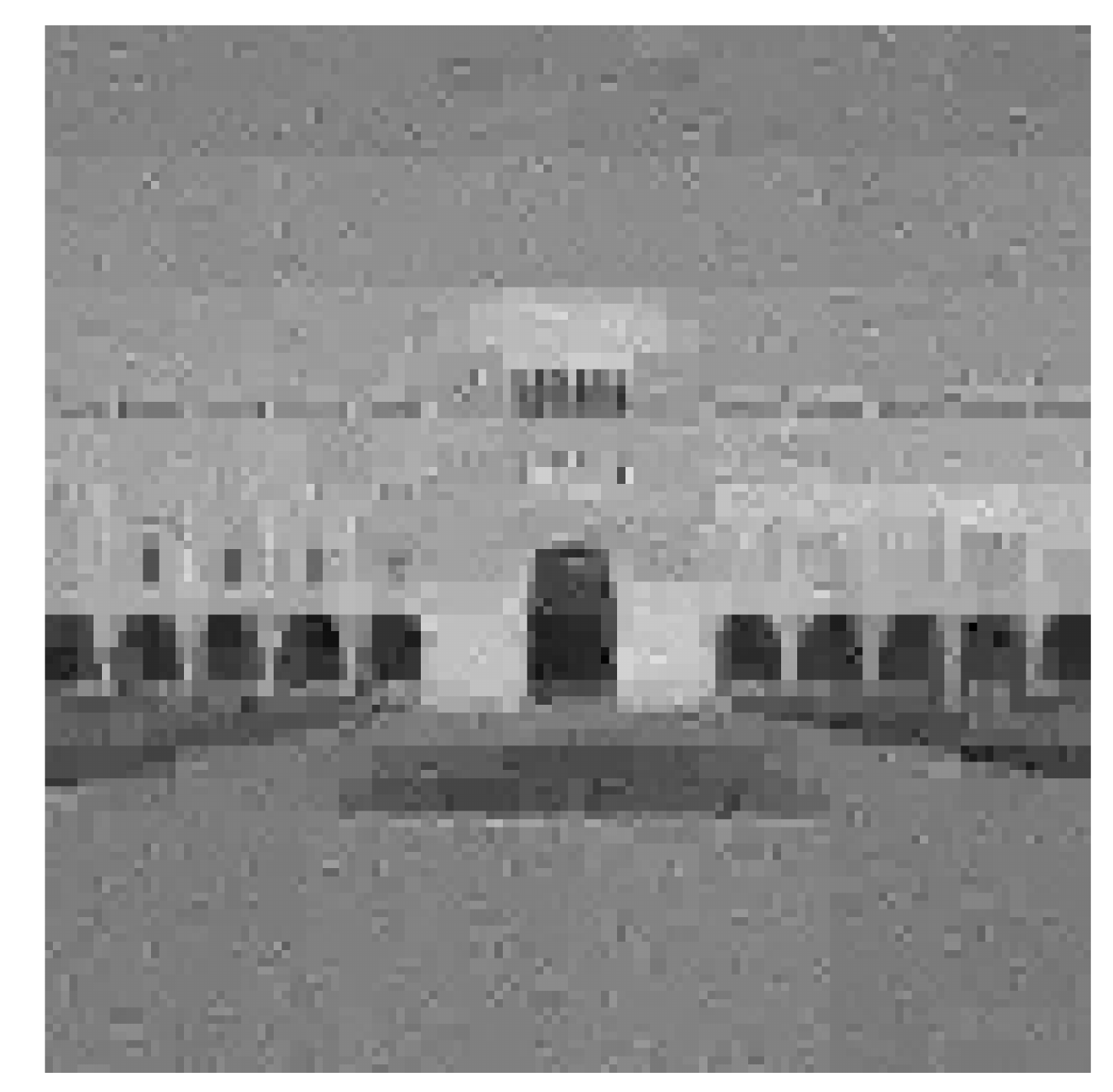} &
			\includegraphics[width=0.22\linewidth]{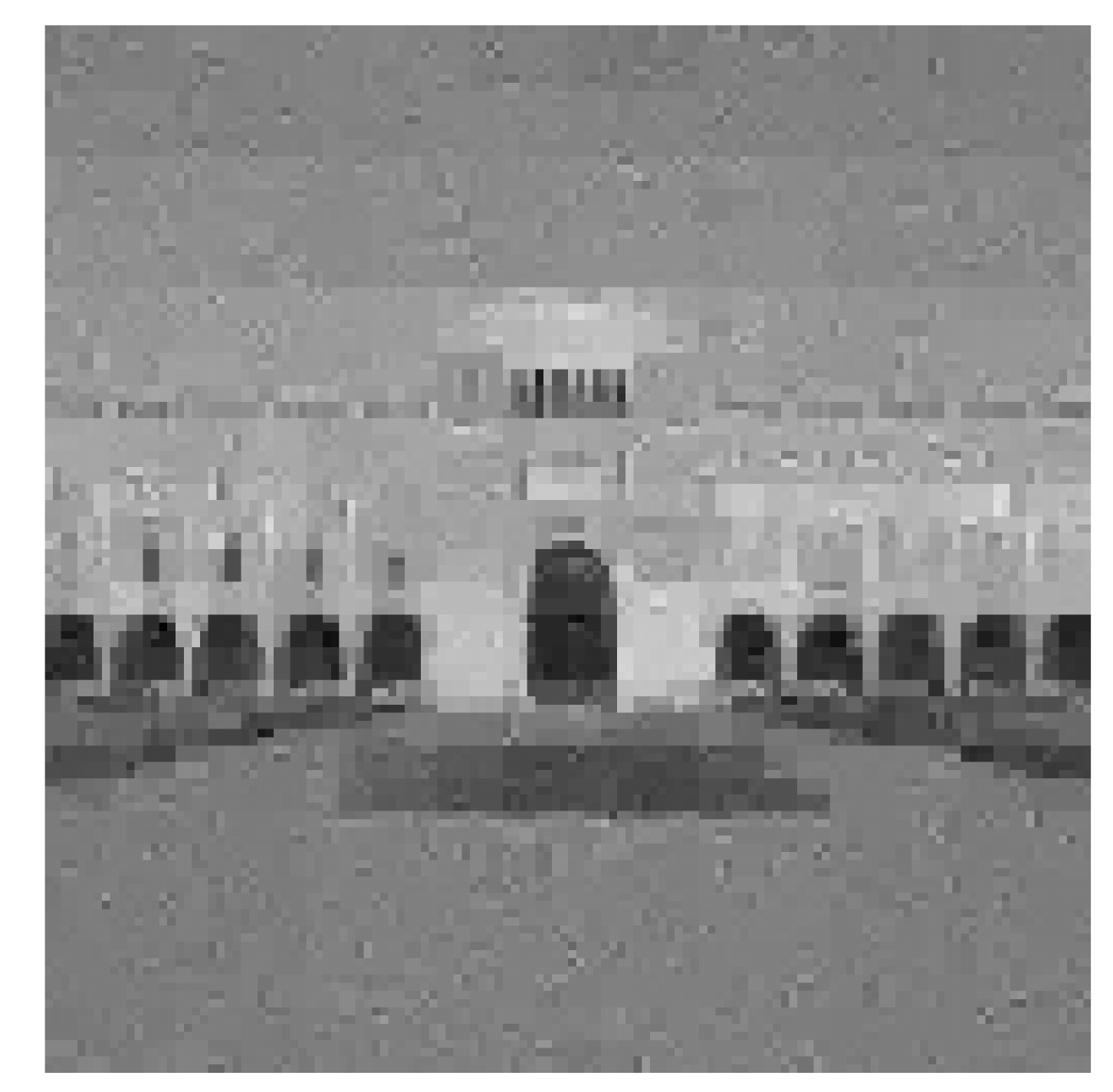}  \\
			& \small{(a) Original image}& \small{(b) $R=4$, SNR $=26.10$dB}& \small{(c) $R=4.25$, SNR $=26.96$dB}& \small{(d) $R=4.5$, SNR $=27.28$dB} \\
			
			\rotatebox{90}{$~~~~~~~m=6000$} &
			\includegraphics[width=0.22\linewidth]{./fig/lovett_original.pdf} &
			\includegraphics[width=0.22\linewidth]{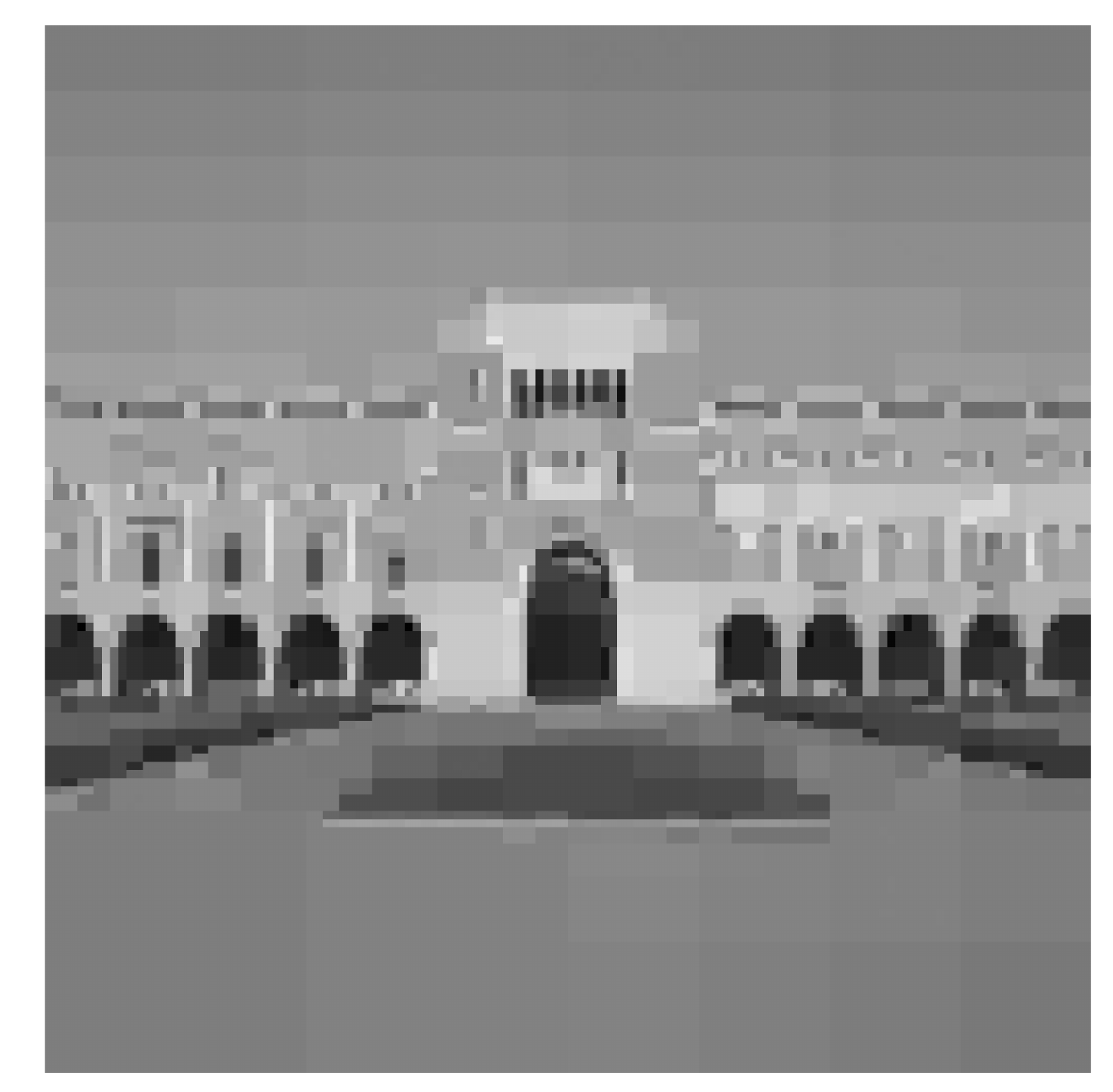} & 
			\includegraphics[width=0.22\linewidth]{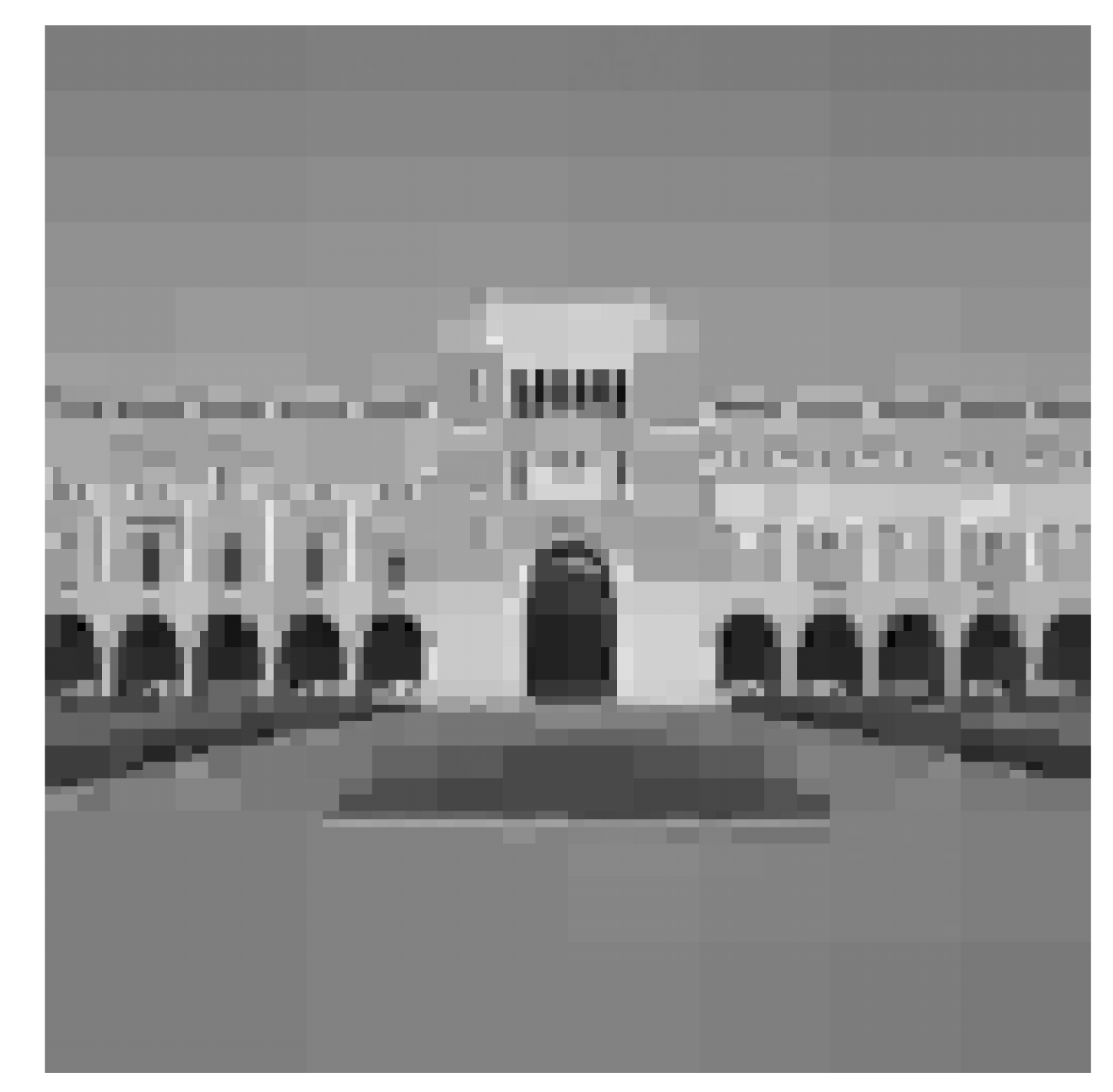} &
			\includegraphics[width=0.22\linewidth]{./fig/r_450_m_6000_s_800_sim.pdf}  \\
			& \small{(a) Original image}& \small{(b) $R=4$, SNR $=79.65$dB}& \small{(c) $R=4.25$, SNR $=81.81$dB}& \small{(d) $R=4.5$, SNR $=83.51$dB} \\
		\end{tabular}
	\end{center}
	\caption{{(a) Original Lovett Hall image ($n=16,384$); sparse reconstructions ($s=800$) using $m=4000$ (top) and $m=6000$ (bottom) measurements for (b) $R=4$, (c) $R=4.25$, (d) $R=4.5$.}}
	\label{fig:lovett}
\end{figure*}

\lemma{Let $\mb{A}$ be the matrix generated as $\mb{A} \sim \mathcal{N}^{m \times n}(0,1)$ and suppose $\mb{x^*, x^0} \in \R^n$ are $s-$sparse vectors satisfying $\norm{\mb{x^* - x^0}}_2 \leq \delta\norm{\mb{x^*}}_2$. Let $\eta \in [0,1],~\eps > 0$. If the number of measurements \\
	$m \geq \frac{2}{\eps^2}\left(s\log{(n)} + 2s\log{\left(\frac{35}{\eps}\right)}+\log{\left(\frac{2}{\eta}\right)}\right)$, then, the following is true with probability exceeding $1 - \eta$:
	
	$$
	d_H(\sgn({\mb{Ax^*}}), \sgn({\mb{Ax^0}})) \leq \frac{\delta}{2} + \eps;
	$$
where $d_H$ is Hamming distance between binary vectors defined as:
$$
d_H(\mb{a,b}) := \frac{1}{n}\sum_{i=1}^{n}a_i \oplus b_i,
$$
for $n-$ dimensional binary vectors $\mb{a,b}$.
\label{lemma1} 
} 

\proof{ 
Given $m \geq \frac{2}{\eps^2}\left(s\log{(n)} + 2s\log{\left(\frac{35}{\eps}\right)}+\log{\left(\frac{2}{\eta}\right)}\right)$, using Theorem 3 from~\cite{Jacques2013} we conclude that $F(\cdot)$ is a B$\eps$SE for $s$-sparse vectors.
Thus for sparse vectors $\mb{x^*, x^0}$:
\begin{align}
\label{eq:th3}
d_H(F\mb{(x^*)},F\mb{(x^0)}) \leq d_S(\mb{x^*,x^0}) + \eps.
\end{align}

Here, $d_S(\cdot)$ is defined as the natural angle formed by two vectors. Specifically, for $\mb{p,q}$ in unit norm ball,
$$
d_s(\mb{p},\mb{q}) := \frac{1}{\pi}\textrm{arccos}\langle\mb{p},\mb{q}\rangle = \frac{1}{\pi}\theta,
$$
where $\theta$ is the angle between two unit norm vectors $\mb{p}$ and $\mb{q}$.

We note that,
$$
\norm{\mb{p-q}}_2 = 2\sin(\frac{\theta}{2}).
$$
Thus,
\begin{align}
\label{eq:dhds}
2d_s(\mb{p},\mb{q}) \leq \norm{\mb{p - q}}_2 \leq \pi d_s(\mb{p},\mb{q})
\end{align}
Combining eq.~\ref{eq:th3} and eq.~\ref{eq:dhds}, we conclude,

\begin{align}
d_H(\sgn({\mb{Ax^*}}), \sgn({\mb{Ax^0}})) &\leq \frac{1}{2}\norm{\mb{x^*-x^0}} + \eps,~~\text{i.e.,}  \nonumber \\
d_H(\sgn({\mb{Ax^*}}), \sgn({\mb{Ax^0}})) &\leq \frac{\delta}{2} + \eps. \label{eq:hd}
\end{align}

}

We use this to obtain:

\theorem{Given an initialization $\mb{x^0}$ satisfying $\norm{\mb{x^* - x^0}}_2 \leq \delta\norm{\mb{x^*}}_2$, for $0 < \delta < 1, \eta \in [0,1],~\eps > 0$, if we have number of (Gaussian) measurements satisfying $m \geq \frac{2}{\eps^2}\left(s\log{(n)} + 2s\log{\left(\frac{35}{\eps}\right)}+\log{\left(\frac{2}{\eta}\right)}\right)$ and $s \leq \gamma m/\left(\log\left(n/m\right)+1\right)$, then the estimate after the first iteration $\mb{x^1}$ of Algorithm \ref{alg:MoRAM} is exactly equal to the true signal $\mathbf{x^*}$ with probability at least $1-K\exp(-cm)-\eta$, with $K$ and $c$ being numerical constants.
}

\proof{In the estimation step, Algorithm \ref{alg:MoRAM} dubs the problem of recovering the true signal $\mb{x^*}$ from the modulo measurements as the special case of signal recovery from sparsely corrupted compressive measurements. As we discussed in Section \ref{sec:modeff}, the presence of modulo operation modifies the compressive measurements by adding a constant noise of the value $R$ in fraction of total measurements. However, once we identify correct bin-index for some of the measurements using $\mb{x^0}$, the remaining noise can be modeled as sparse corruptions $\mb{d}$ according to the formulation:

$$
\mb{y} = \mb{Ax} + \mb{I_nR(p^0-p^*)} = \mb{Ax} + \mb{d}.
$$  
Here, the $\sl{l}0$-norm of $\mb{d}$ gives us the number of noisy measurements in $\mb{y^0_c}$.

If the initial bin-index vector $\mb{p^0}$ is close to the true bin-index vector $\mb{p^*}$, then $\norm{\mb{d}}_0$ is small enough with respect to total number of measurements $m$; thus, $\mb{\mb{d}}$ can be treated as sparse corruption. If we model this corruption as a sparse noise, then we can employ JP for a guaranteed recovery of the true signal given (i) sparsity of the noise is a fraction of total number of measurements; (ii) sufficiently large number of measurements are available.  

We compute $\norm{\mb{d}}_0$ as,

$$
\norm{\mb{d}}_0 =  \norm{(\mb{p^*-p^0})R}_0;
$$

expanding further,
\begin{align*}
\norm{\mb{d}}_0  & = \frac{\mathbf{1}-\sgn(\langle \mathbf{A} \cdot \mathbf{x^0} \rangle)}{2} -  \frac{\mathbf{1}-\sgn(\langle \mathbf{A} \cdot \mathbf{x^*} \rangle)}{2} \\
& = \frac{\sgn({\mb{Ax^*}})- \sgn({\mb{Ax^0}})}{2} \\
& = \frac{F\mb{(x^*)} - F\mb{(x^0)}}{2} \\
& = d_H(F\mb{(x^*)}, F\mb{(x^0)}). \\
\textnormal{From eq.~\ref{eq:hd},} \\
& \leq \frac{\delta}{2} + \epsilon = \gamma m.
\end{align*}

Algorithm \ref{alg:MoRAM} is essentially the Justice Pursuit (JP) formulation as described in \cite{Laska2009}. Exact signal recovery from sparsely corrupted measurements is a well-studied domain with uniform recovery guarantees available in the existing literature. We use the guarantee proved in \cite{li2013compressed} for Gaussian random measurement matrix, which states that one can recover a sparse signal exactly by tractable $\ell_1$-minimization even if a positive fraction of the measurements are arbitrarily corrupted. With $\norm{\mb{d}}_0 \leq \gamma m$, we invoke Theorem $1.1$ from \cite{li2013compressed} to complete the proof.
$\qed$}

\section{Numerical experiments}
\label{sec:exp}

%

In this section, we present the results of simulations of signal reconstruction using our algorithm. All numerical experiments were conducted using MATLAB R2017a on a Linux system with an Intel CPU and 64GB RAM. Our experiments explores the performance of the MoRAM algorithm on both synthetic data as well as real images.

We perform experiments on a synthetic sparse signal $\mb{x^*} \in \R^n$ with $n=1000$. The sparsity level of the signal is chosen in steps of $3$ starting from $3$ with a maximum value of $12$. The non-zero elements of the test signal $\mb{x^*}$ are generated using zero-mean Gaussian distribution $\mathcal{N}(0, 1)$ and normalized such that $\norm{\mb{x^*}} =1$. The elements of the Gaussian measurement matrix $\mb{A} \in \R^{m\times n}, a_{ij}$ are also generated using the standard normal distribution $\mathcal{N}(0, 1)$. The number of measurements $m$ is varied from $m = 100$ to $m=1000$ in steps of $100$.  



Using $\mb{A, x^*}$ and $\R$, We first obtain the compressed modulo measurements $\mb{y}$ by passing the signal through forward model described by Eq.~\ref{eq:modmeas2}. We compute the initial estimate $\mathbf{x^0}$ using the algorithm~\ref{alg:RCM}. For reconstruction, algorithm~\ref{alg:MoRAM} is employed. we plot the variation of the relative reconstruction error ($\frac{\norm{\mathbf{x^*-x^T}}}{\norm{\mathbf{x^*}}}$) with number of measurements $m$ for our AltMin based sparse recovery algorithm MoRAM.

For each combination of $R, m$ and $s$, we run $10$ independent Monte Carlo trials, and calculate mean of the relative reconstruction error over these trials. Fig.~\ref{fig:plot} (a), (b) and (c) illustrate the performance of our algorithm for increasing values of $R$ respectively. It is evident that for each combination of $R$ and $s$, our algorithm converges with probability $1$ to give the exact recovery of the true signal (zero relative error) provided enough number of measurements. In all such cases, the minimum number of measurements required for exact recovery are well below the ambient dimension $(n)$ of the underlying signal. 

\subsection{Experiments on real image}
We also evaluated the performance of our algorithm on a real image. We obtain sparse representation of the real image by transforming the original image in the wavelet basis (db1). The image used in our experiment is $128 \times 128$ ($n=16384$) image of Lovett Hall (fig.~\ref{fig:lovett}(a)), and  we use the thresholded wavelet transform (with Haar wavelet) to sparsify this image with $s = 800$. We reconstruct the image with MoRAM using $m = 4000$ and $m=6000$ compressed modulo measurements, for $3$ different values of $R$, $4,4.25$ and $4.5$. As expected, the reconstruction performance increases with increasing value of $R$. As shown in Fig.~\ref{fig:lovett}(bottom), for $m=6000$, The algorithm produces near-perfect recovery for all $3$ values of $R$ with high PSNR.

\section{Discussion}
\label{sec:disc}
In this paper, for signal recovery from compressed modulo measurements, we presented a novel algorithmic approach inspired from the classical phase retrieval solutions. Our mathematical and experimental analysis support our claim of exact signal recovery through proposed algorithm. Several open questions remain that can serve as the future directions of our work. While in this paper we considered only two periods within the modulo operation, extending the proposed approach for more periods (and theoretically infinite periods) is a significant and interesting research direction. Instead of relying on sparsity prior for compressed recovery, employing novel set of priors such as GAN priors\cite{bora2017compressed,shah2018solving} can also be a direction to be explored. Analysis and guaratees for initialization in our algorithm can also be an interesting direction. Moreover, our analysis is limited to the case of Gaussian measurements, thus extending our results to various measurement schemes such as Fourier samples can be an interesting problem for future study.
\bibliographystyle{unsrt}
\bibliography{./vsbib}

\end{document}